\documentclass{article}

\usepackage{arxiv}

\usepackage[utf8]{inputenc} 
\usepackage[T1]{fontenc}    
\usepackage{hyperref}       
\usepackage{url}            
\usepackage{booktabs}       
\usepackage{amsfonts}       
\usepackage{nicefrac}       
\usepackage{microtype}      
\usepackage{lipsum}		
\usepackage{graphicx}
\usepackage{natbib}
\usepackage{doi}
\usepackage{amsmath}
\usepackage{multirow}

\title{TaskDrop: A Competitive Baseline for Continual Learning of Sentiment Classification}

\author{ 
	{Jianping Mei}\thanks{The authors contribute equally to this paper.} \\
	Zhejiang University of Technology\\
	\texttt{jpmei@zjut.edu.cn} \\
	\And
	{Yilun Zheng}\footnotemark[1] \\
	Zhejiang University of Technology\\
	\texttt{zhengyilunscholar@foxmail.com} \\
	\AND
	Qianwei Zhou \\
	Zhejiang University of Technology \\
	\texttt{zhouqianweischolar@gmail.com} \\
	\And
	Rui Yan \\
	Zhejiang University of Technology \\
	\texttt{Ryan@zjut.edu.cn} \\
}

\renewcommand{\shorttitle}{\textit{arXiv} Template}

\hypersetup{
pdftitle={A template for the arxiv style},
pdfsubject={q-bio.NC, q-bio.QM},
pdfauthor={David S.~Hippocampus, Elias D.~Striatum},
pdfkeywords={First keyword, Second keyword, More},
}

\begin{document}
\maketitle
\begin{abstract}
	In this paper, we study the multi-task sentiment classification problem in the continual learning setting, i.e., a model is sequentially trained to classifier the sentiment of reviews of products in a particular category. The use of common sentiment words in reviews of different product categories leads to large cross-task similarity, which differentiates it from continual learning in other domains. This knowledge sharing nature renders forgetting reduction focused approaches less effective for the problem under consideration. Unlike existing approaches, where task-specific masks are learned with specifically presumed training objectives,
	we propose an approach called Task-aware Dropout (TaskDrop) to generate masks in a random way. While the standard dropout generates and applies random masks for each training instance per epoch for effective regularization, TaskDrop applies random masking for task-wise capacity allocation and reuse. We conducted experimental studies on three multi-task review datasets and made comparison to various baselines and state-of-the-art approaches. Our empirical results show that regardless of simplicity, TaskDrop overall achieved competitive performances for all the three datasets, especially after relative long term learning. This demonstrates that the proposed random capacity allocation mechanism works well for continual sentiment classification.
\end{abstract}

\section{Introduction}
The capability of learning new tasks while maintaining performance on learned ones is required for many practical applications where tasks are learned sequentially. For example, retraining a robot or a sentiment classification system each time encountering a new task is cumbersome or even impossible when previous data are no longer accessible.
The sequential learning or continual learning ability is also fundamental for advanced artificial intelligence systems to adapt to unknown tasks \cite{Legg2007,thrun1995lifelong}. 
When sequentially learning tasks with very limited or no access to data of previous tasks, the model tends to forget what has been learned, leading to degraded performance on previous tasks \cite{CF_McCloskey,CF_Ratcliff}.

Most of the recent efforts in continual learning are made to deal with this so-called Catastrophic Forgetting problem for learning with deep neural network models.  
Some of them follow the joint training idea by replying some forms of previous information that are stored \cite{icarl,GEN,AGEN2019,2019Experience,han-etal-2020} or synthesized with a generative model \cite{shin2017continual,nguyen2018variational} when learning new tasks. Others focus on preserving the learned model by strictly freezing or penalizing large changes to the subset of model parameters which are regarded to be important to previous tasks \cite{french1991using,he2018CAB,Masse2018,PackNet,PNN,PathNet,EWC,2019Learn,RandomPath2019}.

Although freezing-based approaches work in a more direct way for model preservation and hence forgetting reduction than regularization-based ones, the constant demand on additional capacity for new tasks becomes an issue when the number of tasks is large.  In this work, we focus on the continual learning of sentiment classification tasks with a size-fixed encoder without re-accessing to data of previous tasks.

Given the overall model capacity, preserving existing knowledge inevitably restricts the learning of new tasks.
Facing the tradeoff between remembering what have been learned and exploring new knowledge with a risk of forgetting, it is reasonable to favour the former if these tasks have low transferbility because of their difference in distribution and other aspects, such as classification of different image datasets \cite{LwF,HAT} or learning different types of relations \cite{han-etal-2020}. For the sentiment classification scenario, typical sentiments and words used to express these sentiments are quite similar across tasks, e.g., ``amazing'' is a commonly used positive word when one makes comments on books or anything else. This high cross-task similarity nature increases the chance for backward transfer, i.e., learning from new tasks helps to strengthen the common knowledge base shared by previous tasks. For such kind of problems, preserving-oriented approaches usually become less effective as observed in \cite{KAN} as well as in our experiments.  

Instead of completely avoid reactivation of those preserved parameters, masking-based approaches learn non-exclusive masks so that frozen units still have the opportunity to be re-activated \cite{PathNet,CAT,KAN}. 
A key problem for masking-based approaches is the way to obtain masks. Existing approaches learn masks based on different objectives, which are designed for specific type of problems. In this paper, we investigate a simple randomization-based approach called Task-aware Dropout (TaskDrop). At the beginning of a task, we generate random binary vectors as masks for each of the layers, and apply the corresponding mask to units of each layer during forward and backward passes. This random unit masking operation is also adopted in the well known dropout, an important trick for deep learning. While the standard dropout applies random masks to sample a large number of subnetworks for each forward pass without considering task boundaries, masking in TaskDrop works as random capacity allocation  for each coming task. Capacity reuse or re-activation across tasks is simply controlled by the dropout rate, rather than guided by any presumed objectives as existing approaches. This gives the flexibility for TaskDrop to adjust its conservativeness to a proper degree, leading to better adaptiveness to different problems.

We carried out sentiment classification experiments in the continual learning setting on three multi-task datasets with different levels of transfer accuracy. Each task performs sentiment polarity prediction of reviews of products from one category. 
Along with state-of-the-art approaches, we also include the results of four baselines to account for various naive solutions as well as an upper-bound solution. Main contributions of this work are summarized as below:
\begin{itemize}
	\item We investigate a simple masking-based approach for sequential learning of sentiment classification tasks. Masking with randomly generated task specific masks  results in random capacity reuse, which is controlled by the dropout rate. 
	\item Analytical discussions and empirical results show that the random capacity reuse mechanism works competitively compared to other elaborately designed approaches.
	\item We carried out experimental study on the robustness of continual learning approaches with tasks of different levels of relevance, to investigate how they perform when the nature of the target problem does not well fit their assumptions. 
	
\end{itemize}

\section{Continual Learning with Random Task Masks}

\subsection{Masking-based Continual Classification}
\begin{table*}
	\centering
	\resizebox{\textwidth}{!}{ 
		\begin{tabular}{lccccc}
			\toprule
			Method & Fixed model size & Way for mask obtaining & Mask sparsity control & Mask overlap\\ \hline
			PackNet \cite{PackNet} & no & pruning with assigned ratio & pruning ratio& no\\
			HAT \cite{HAT} & yes & minimizing the weighted $l_1$ norm of attentions& regularization weight $c$ &yes\\
			KAN \cite{KAN}& yes & minimizing the training loss of the current task& no direct control&yes\\
			TaskDrop & yes& random sampling &retention ratio $p$ & yes\\
			\bottomrule
		\end{tabular}
	}
	\caption{Comparison of characteristics of masking-based continual learning approaches.}
	\label{tb:mask_char}
\end{table*}
Masking-based approaches apply masks to parameters or unit activations to decide what to access and to update when learning a new task.  
The most important difference among these approaches is thus how these masks are generated or learned. To preserve existing knowledge, PackNet \cite{PackNet}
generates exclusive masks with an assigned pruning ratio to directly freeze large-valued parameters, while HAT \cite{HAT}
only decreases the chance for reactivating units that have been already activated in previous tasks. In the most related approach KAN \cite{KAN}, which also works on continual sentiment classification, masks are learned to activate units to give optimized learning of the current task.

Inspired by the success of the dropout technique for deep learning, we explore a new solution based on random task-aware masks for the problem of continual learning. Instead of treating masks as additional parameters to be learned simultaneously with the network, our approach called Taskdrop directly generates masks like PackNet, but allows capacity reuse like those learnable masks. Table \ref{tb:mask_char} compares characteristics of the proposed  masking-based approach with those of existing ones. Next, we present the details on instantiating this idea into a solution for continual sentiment classification.

\begin{figure}
	\centering
	\includegraphics[width=0.8\textwidth]{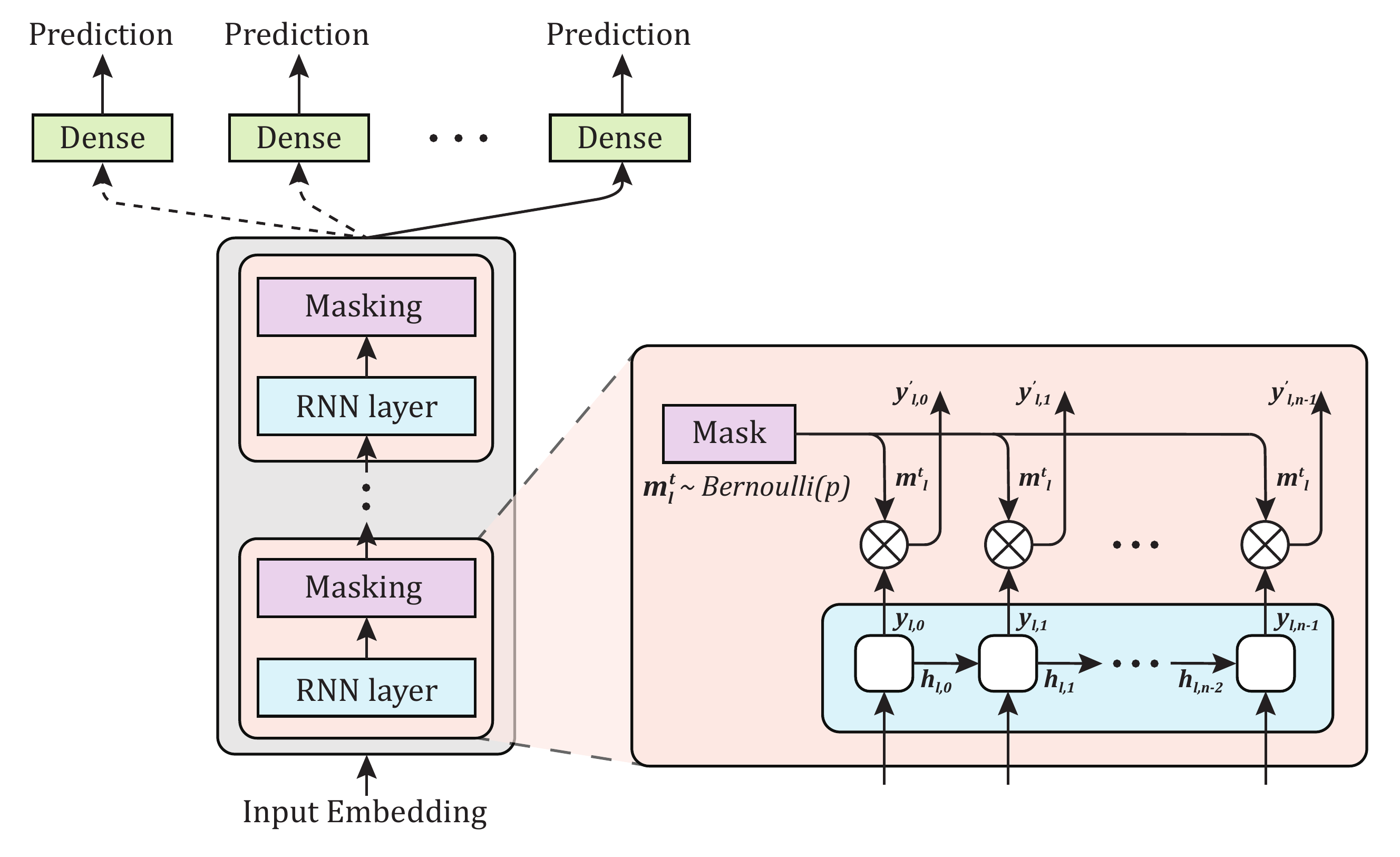}
	\caption{An overall model structure of TaskDrop consisting of a RNN encoder and multi-head dense layers. Each RNN layer is followed by the task-aware random masking operation with its details on the right. }
	\label{fig:model}
\end{figure}

\subsection{Task-Aware Dropout} 
\textbf{Overview.}
The overall structure of TaskDrop is illustrated in Figure \ref{fig:model}. It consists of a RNN backbone with a multi-MLP output layer, where each MLP corresponds to the classifier of a certain task. It is pretty much the same as the main network in KAN \cite{KAN}. While in KAN an individual accessibility module, i.e., another RNN is trained in an alternating way with the main network for learning the masks, here in TaskDrop we only have the main network to learn.

Assume that the input embedding for the $j$th training case of task $t$, i.e., a piece of text with fixed length $n$ is $\mathbf{X}^t_j=[\mathbf{x}_0, \mathbf{x}_1, \ldots, \mathbf{x}_n]$, and the output of the $l$th hidden layer at timestep $i$ is $\mathbf{y}_{l,i}$. For each hidden layer $l$ with $n_l$ units, a binary mask $\mathbf{m}^t_l \in \{0,1\}^{n_{l}}$ is element-wise multiplied to each of the $n$ timestep outputs of this layer. 
At the beginning of a task, we generate masks for all the layers $\{\mathbf{m}^t_l\}_l$ by randomly and independently drawing each element from the Bernouli distribution with a probability of $p$ for being 1, i.e.,
\begin{equation}
	\label{eq:bernoulli}
	\mathbf{m}^t_l=[Bernoulli(p),...,Bernoulli(p)].
\end{equation}
These masks are stored and retrieved in every forward and backward pass during the whole learning process of this task. 
The probability $p$ decides how likely that a hidden layer unit is retained or activated for this task. Simply speaking, $p$ controls the sparsity of activated units of each layer, which is formulated with the $l_1$ norm of parameter \cite{yoon2018DEN} or attention \cite{HAT} of each layer. Increasing the value of $p$ is likely leading to a model with more extensive unit sharing, or reuse.

After getting the class prediction, a cross-entropy loss is calculated for updating parameters of RNN and the $t$th classifier. 
Next, we elaborate more on forward passes and back-propagations with masking applied RNN.

\textbf{Masking of RNN.} 
Following the work on modified dropout to RNN structures \cite{zaremba2014recurrent}, the masking operation as illustrated in Figure \ref{fig:model} only works on non-recurrent connections denoted by dashed arrows. 
To be more clear, given $\mathbf{m}^t_l$ and the RNN output $\mathbf{y}_{l,i}$=$\mathbf{h}_{l,i}$ at timestep $i$ for a training case of task $t$, where $\mathbf{h}_{l,i}$ is the recurrent state at this timestep, we apply the masking operation to $\mathbf{y}_{l,i}$ only as below during feed-forward
\begin{align}
	\mathbf{y}'_{l,i} =  \mathbf{y}_{l,i}\odot \mathbf{m}^t_l.
\end{align}
For recurrent connections as shown by solid arrows, information still flow through the units even they are masked out, so that the valuable memorization ability of GRU is not sacrificed. 
During back propagation, masked units receive no gradient. 

Another detail is that we applied unit masking like the standard dropout, as well as HAT and KAN, while PackNet operates masking over weights.

\subsection{Why Random Masks Work?}

We now provide some analytical discussions on unique properties of TaskDrop to hopefully shed some light on understanding and explanation of the underlying working principle of this simple approach.

\begin{figure}
	\centering
	\includegraphics[width=0.75\textwidth]{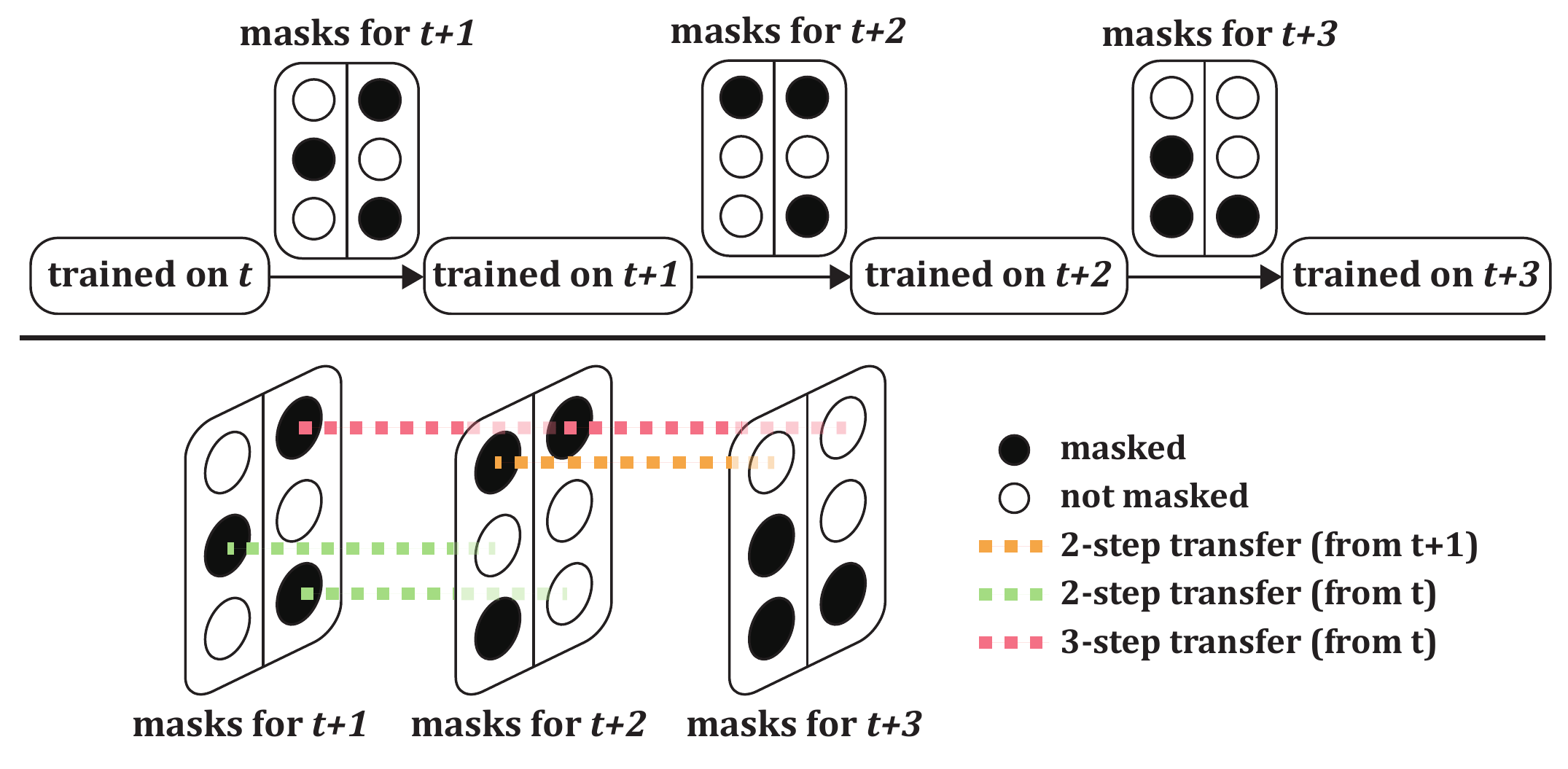}
	\caption{Illustration of skip-task transfer. Top: Masking-based learning of tasks in sequence. Bottom: different skip-task transfers produced by random masks of three subsequent tasks.}
	\label{fg:skip_task}
\end{figure}

\textbf{Skip-task transfer. \label{sec:skip}}
The random masking operation gives TaskDrop the ability of $s$-step skip-task transfer with $s > 1$ as illustrated in Figure\ref{fg:skip_task}. Specifically, parameters of the current task $t$ may be preserved and used by the $s$th subsequent task with $s\geq 2$ if the corresponding units happen to be masked during all the next $s-1$ tasks, i.e., a preservation duration of $s-1$ tasks. This mechanism allows direct knowledge transfer from one task to another that is $s$ steps way on the task stream.  

\begin{figure}
	\centering
	\includegraphics[width=0.75\textwidth]{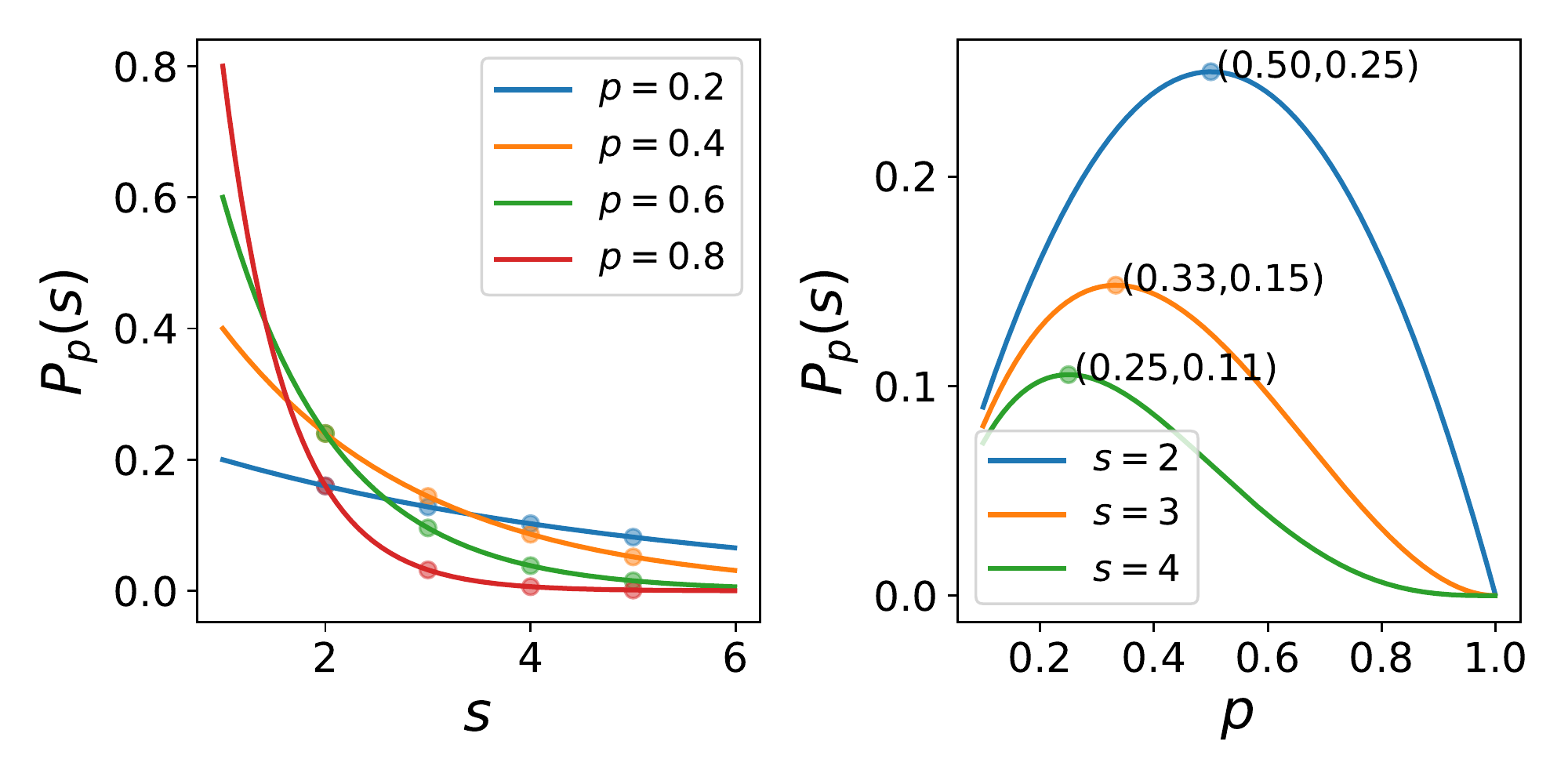}
	\caption{The probability of $s$-step skip-task transfer for each individual unit $P_p(s)$ with respect to $s$ on the left and with respect to retention ratio $p$ on the right. }
	\label{fg:skip_p}
\end{figure}

Given $p$ the probability of being not masked, the probability of $s$-step sharing for each individual unit is $P_p(s)=(1-p)^{s-1}p$. Figure \ref{fg:skip_p} plots the curves of $P_p(s)$ with respect to $s$ for $p=\{0.2,0.4,0.6,0.8\}$ on the left and with respect to $p$ for $s=\{2,3,4\}$ on the right. As shown from the left plot of this figure, $P_p(s)$ is monotonically decreasing with respect to $s$, and the larger the $p$ is, the faster it decreases from with the initial value that equals to $p$. Let us focus on the conditions that lead to large $P_p(s)$ with $s \geq 2$ on the right. It is clear that $s=2$ has the largest probability for all $p \in (0,1)$, and when $p$ is either too small or too large, i.e., $0.3\leq p \leq 0.7$, we have $P_p(s=2) \geq 0.2$. If increasing the steps to $s=3$, the largest $P_p(s=3)$ is 0.15 when $p$ is around 0.3. 
Although probabilities of skip-task transfers for $s \geq 2$ are relative small compared to direct transfer, that is, re-activate in the next task, they actually help TaskDrop to  achieve better results as demonstrated with empirical results.

Based on the above analysis that connects random masks with skip-task knowledge sharing, we next give some discussions by theoretical comparison between TaskDrop and other closely related approaches to further discuss the merits of random masks for continual learning.

\textbf{Parameter sharing: whole vs. portional.}
When $p\rightarrow 1$, TaskDrop reduces to the No-masking baseline, which shares all the encoder parameters in learning the next task. In other words, No-masking is a hundred percent reuse variant of TaskDrop with no skip-task sharing. It may give good results if the next task is highly similar to the current one, which means a smooth task boundary. Even though, masking a relative small portion of weights to allow some skip-task sharing makes the model more robust and usually performs better in relative long term learning as observed in our experiments. 
In cases that tasks are of little similarities, a large portion of parameter sharing could cause severe forgetting. For TaskDrop, we can adjust it to give more conservative solutions with a smaller $p$, leading to reduced activation overlap among representations \cite{french1991using}.

\textbf{Preserving: temporary vs. permanent.  }
Freezing-based approaches permanently preserve parameters by prohibiting re-activation in all subsequent task learning. The $s$-step skip-task transfer in TaskDrop works as temporary preserving for a period of $s-1$ task learning. Although no-reactivation effectively solves the catastrophic forgetting problem, fresh capacity is constantly required in order to learn new tasks. That is to say, for a fixed model size, such kind of one-time using exhausts all the capacity after a certain number of tasks. Moreover, being over conservative results in unsatisfied results for the problem under consideration, where knowledge sharing is important.  

\textbf{Dropout: per-sample each iteration vs. per-task.}
Regardless the similarity in the way of mask generation and application, masks used in dropout and TaskDrop are different in both their lasting time and the number of instances to be applied for. The standard dropout generates random masks for each training sample each time it is processed, and the sampled network is learned upon a single instance. In TaskDrop, masks are generated at a per-task frequency and shared by all the samples during the whole learning process of this task. The above differences indicate that while dropout is an effective trick for improving model generalization ability, it is not designed for continual learning as task boundaries are unaware during the entire learning process. 

\subsection{Limitations}	
Since the skip-task transfer mechanism works properly after learning an enough number of tasks, TaskDrop may not be a good choice for learning  only a very short task sequence. For example, when sequentially learning two tasks, random masks for the second task have little change of being better than those learned ones. Considering that TaskDrop incurs no additional parameters to be learned, we should have resources left for adopting a higher capacity classifier as a remedy to the effectiveness gap between random and learned masks if there is any. 

\section{Related Work}
Given the long and diverse literature of continual learning, we only focus on recent approaches with a similar problem setting, i.e., continual task learning with neural network models without replaying data of previous tasks. Since dropout itself is not the focus of this paper, works on different dropout techniques are also excluded here.

\textbf{Regularization-based methods.} Structural regularization approaches penalize changes to learned knowledge when learning a new task \cite{EWC,IMM, LwF,zenke2017SI,Aljundi2018,he2018CAB,wang-etal-2019,yoon2018DEN}. These approaches use different representations of the existing knowledge that they attempt to preserve, including output predictions \cite{LwF}, hidden spaces \cite{TrikiABT17encoder}, or model parameters \cite{EWC,yoon2018DEN}. The parameter importance may be measured with the diagonal of the Fisher information matrix \cite{EWC} or based on the sensitivity of
the learned function to their changes after convergence \cite{Aljundi2018}, or computed during training in an online manner \cite{zenke2017SI}. The work in \cite{IMM} further extends \cite{EWC} with a separate model-merging step after learning a new task. A recent approach uses both regularization and memory-based example replay \cite{2021Huang}.

\textbf{Freezing-based approaches.}
Although regularization-based approaches work with well-defined objectives under certain assumptions, they only take implicit and indirect control over the forgetting problem.
A straightforward way to overcome catastrophic forgetting is to immediately freeze parameters learned for a task and seek for additional capacity for new task learning \cite{PNN,PathNet,PackNet,yoon2018DEN}. 
Some approaches pre-allocate certain capacity \cite{PNN} or dynamically add capacity for a coming task \cite{yoon2018DEN}, leading to models with a growing size. On the contrary, 
\cite{PackNet} starts with a size-fixed model, and frees part of capacity through parameter pruning. However, capacity distribution over tasks based on scheduled pruning ratios is not applicable when the number of tasks is unknown in advance. Moreover, allowing no reuse at all makes the model stop learning when the capacity limitation is reached.

\textbf{Masking-based approaches.}
Masking-based approaches use masks for selective capacity reuse. Masks are learned to optimize specific objectives. In \cite{HAT}, masks with small sparsity and attention overlap are learned simultaneously with the network training by minimizing the weighted $l_1$ norm of hard attentions. In the approach that is specifically designed for sentiment classification \cite{KAN},  masks are learned with an additional so-called accessibility module in an alternating way with the main network to optimize the performance of the current task. The work \cite{RandomPath2019} is an improvement
over \cite{PathNet} for encouraging knowledge sharing and reuse.

In order to be conveniently trained with the rest of network using SGD, masks are relaxed to continuous attentions with a sigmoid function, which are gradually harden and become binary after an annealing process in \cite{HAT}. Several later works follow this annealing strategy as well as masked gradient for training masks \cite{2020iTAML, KAN,CAT}. In a following work of \cite{KAN}, separate steps to measure whether two tasks are similar or not by comparing to  reference models \cite{CAT}. 

\begin{table*}
	\centering
	\resizebox{\textwidth}{!}{ 
		\begin{tabular}{cccccc|cc}
			\toprule
			Metric&	Dataset & Individual Networks & Classify-only &  No-masking & TaskDrop & Multi-task & \small  \\
			\midrule
			\multirow{4}{*}{$A^{\leq 2}$} 
			&high-6 & 82.73 $\pm$ 1.41 & 84.03 $\pm$ 1.41 & \textbf{ 86.03 $\pm$ 1.54} & 84.97 $\pm$ 1.38 & 86.75 $\pm$  1.36 \\
			&mix-24 & 79.12 $\pm$ 2.90 & 80.71 $\pm$ 6.05 & \textbf{ 82.90 $\pm$ 3.20} & 81.23 $\pm$ 3.16 & 84.13 $\pm$ 2.78 \\
			&low-6 & 74.33 $\pm$ 3.94 & 73.29 $\pm$ 4.53 & 78.84 $\pm$ 2.59 & \textbf{ 80.04 $\pm$ 3.64} & 82.12 $\pm$ 2.77 \\\cline{2-7}
			&Average of three
			& 78.73 & 79.34 & \textbf{ 82.59} & 82.08 & 84.33 \\
			\midrule
			\multirow{4}{*}{$\rho^{\leq 2}$}
			&high-6 & -19.00 $\pm$ 4.15 & -15.33 $\pm$ 4.56 & \textbf{ -10.64 $\pm$ 6.24} & -13.30 $\pm$ 3.78 & 00.00 \\
			&mix-24 & -10.57 $\pm$ 8.03 & -12.02 $\pm$ 16.56 & \textbf{ -3.70 $\pm$ 3.92} & -8.54 $\pm$ 3.49 & 00.00 \\
			&low-6 & -50.44 $\pm$ 30.99 & -52.14 $\pm$ 27.86 & -33.54 $\pm$ 17.99 & \textbf{ -26.79 $\pm$ 10.74} & 00.00 \\\cline{2-7}
			&Average of three
			& -26.67 & -26.50 & \textbf{ -15.96} & -16.21 & 00.00 \\
			\midrule
			\multirow{4}{*}{$A^{\leq T}$}
			&high-6 & 81.88 $\pm$ 0.55 & 83.58 $\pm$ 0.58 & 87.47 $\pm$ 0.89 & \textbf{ 87.86 $\pm$ 0.85} & 90.83  \\
			&mix-24 & 78.33 $\pm$ 1.13 & 78.26 $\pm$ 3.72 & 87.47 $\pm$ 1.03 & \textbf{ 87.87 $\pm$ 0.82} & 90.44  \\
			&low-6 & 72.37 $\pm$ 2.74 & 71.05 $\pm$ 4.54 & 79.06 $\pm$ 2.07 & \textbf{ 80.83 $\pm$ 0.88} & 87.16  \\\cline{2-7}
			&Average of three
			& 77.53 & 77.63 & 84.67 & \textbf{ 85.52} & 89.48 \\
			\midrule
			\multirow{4}{*}{$\rho^{\leq T}$}
			&high-6 & -21.72 $\pm$ 1.28 & -17.49 $\pm$ 1.42 & -8.02 $\pm$ 2.19 & \textbf{ -7.06 $\pm$ 2.05} & 00.00 \\
			&mix-24 & -32.44 $\pm$ 3.24 & -33.06 $\pm$ 10.17 & -8.01 $\pm$ 3.25 & \textbf{ -6.76 $\pm$ 2.19} & 00.00 \\
			&low-6 & -52.74 $\pm$ 9.24 & -55.81 $\pm$ 15.61 & -30.70 $\pm$ 9.59 & \textbf{ -23.56 $\pm$ 3.44} & 00.00 \\\cline{2-7}
			&Average of three
			& -35.63 & -35.45 & -15.57 & \textbf{ -12.46} & 00.00 \\
			\bottomrule
		\end{tabular}
	}
	\caption{Comparison with reference approaches in averaged accuracy $A^{\leq t}$ and forgetting ratio $\rho^{\leq t}$ for two and all tasks.}
	\label{Table:ave_three_ref}
\end{table*}
\section{Experiments}
We provide empirical results on three different subsets of review data to evaluate the performance of TaskDrop for continual sentiment classification. We compare TaskDrop with different reference and state-of-the-art approaches for their performance after short and long term sequential learning and visualization of learned representations. Comparison is also made between TaskDrop and the standard dropout. Finally, we study the impact of hyper-parameter. More details on experimental settings and additional results are included in the technical appendix due to space limitation.

\subsection{Setups}

\textbf{Data.} 
We use the continual sentiment classification data from \cite{KAN}. This dataset consists of Amazon reviews of 24 different categories of products, and each category makes up a task. In order to carry out more comprehensive evaluations on how each approach performs on tasks with different representational relevance, we further extract another two subsets with six categories each from this twenty-four dataset. Specifically, we define the Mutual Transfer Accuracy (MTA) between two tasks as the average of testing accuracy on one task of the model trained on the other, and then select six tasks that have the largest or smallest total MTA to form high-6 and low-6, respectively. The original dataset that has a mixed MTA level is referred as to mix-24. 

\begin{table*}
	\centering
	\resizebox{\textwidth}{!}{ 
		\begin{tabular}{cccccccccc}
			\toprule
			Metric&Dataset &  LwF & EWC &HAT&KAN  & TaskDrop  \small  \\
			\midrule
			\multirow{4}{*}{$A^{\leq 2}$} 
			&high-6 & \textbf{ 85.38 $\pm$ 1.35} & 83.78 $\pm$ 1.83 & 84.68 $\pm$ 1.69 & 85.26 $\pm$ 1.79 & 84.97 $\pm$ 1.38 \\
			&mix-24 & 80.74 $\pm$ 4.94 & \textbf{ 82.13 $\pm$ 2.65} & 80.75 $\pm$ 3.04 & 81.39 $\pm$ 2.64 & 81.23 $\pm$ 3.16 \\
			&low-6 & 76.25 $\pm$ 3.36 & 79.74 $\pm$ 2.82 & \textbf{ 80.38 $\pm$ 3.19} & 80.34 $\pm$ 3.05 & 80.04 $\pm$ 3.64 \\\cline{2-7}
			&Average of three
			& 80.79 & 81.88 & 81.94 & \textbf{ 82.33} & 82.08 \\
			\midrule
			\multirow{4}{*}{$\rho^{\leq 2}$}
			&high-6 & \textbf{ -11.71 $\pm$ 4.70} & -15.88 $\pm$ 3.48 & -13.57 $\pm$ 4.01 & -12.09 $\pm$ 4.67 & -13.30 $\pm$ 3.78 \\
			&mix-24 & -7.59 $\pm$ 9.10 & -3.79 $\pm$ 13.58 & -6.21 $\pm$ 4.79 & \textbf{ -3.76 $\pm$ 4.53} & -8.54 $\pm$ 3.49 \\
			&low-6 & -44.07 $\pm$ 25.18 & -28.76 $\pm$ 13.44 & \textbf{ -25.92 $\pm$ 11.30} & -27.30 $\pm$ 15.41 & -26.79 $\pm$ 10.74 \\\cline{2-7}
			&Average of three
			& -21.12 & -16.14 & -15.23 & \textbf{ -14.38} & -16.21 \\
			\midrule
			\multirow{4}{*}{$A^{\leq T}$}
			&high-6 & 86.95 $\pm$ 0.65 & 85.40 $\pm$ 0.42 & 85.86 $\pm$ 0.45 & 87.19 $\pm$ 0.70 & \textbf{ 87.86 $\pm$ 0.85} \\
			&mix-24 & 83.38 $\pm$ 3.18 & 86.59 $\pm$ 0.65 & 82.99 $\pm$ 0.36 & 84.39 $\pm$ 1.04 & \textbf{ 87.87 $\pm$ 0.82} \\
			&low-6 & 76.70 $\pm$ 2.29 & 77.75 $\pm$ 3.69 & 78.97 $\pm$ 1.39 & 79.27 $\pm$ 1.54 & \textbf{ 80.83 $\pm$ 0.88} \\ \cline{2-7}
			&Average of three
			& 82.34 & 83.25 & 82.61 & 83.62 & \textbf{ 85.52} \\
			\midrule
			\multirow{4}{*}{$\rho^{\leq T}$}
			&high-6 & -9.31 $\pm$ 1.59 & -13.01 $\pm$ 1.09 & -12.04 $\pm$ 1.16 & -8.75 $\pm$ 1.67 & \textbf{ -7.06 $\pm$ 2.05} \\
			&mix-24 & -18.51 $\pm$ 8.36 & -9.91 $\pm$ 1.73 & -19.65 $\pm$ 1.15 & -16.32 $\pm$ 3.11 & \textbf{ -6.76 $\pm$ 2.19} \\
			&low-6 & -38.30 $\pm$ 7.65 & -35.11 $\pm$ 14.59 & -28.85 $\pm$ 6.13 & -30.57 $\pm$ 6.85 & \textbf{ -23.56 $\pm$ 3.44} \\\cline{2-7}
			&Average of three
			& -22.04 & -19.34 & -20.18 & -18.55 & \textbf{ -12.46} \\
			\bottomrule
		\end{tabular}
	}
	\caption{Comparison with state-of-the-art approaches in averaged accuracy $A^{\leq t}$ and forgetting ratio $\rho^{\leq t}$ for two and all tasks}
	\label{Table:ave_three_SOTA}
	
\end{table*} 

\textbf{Baselines} We consider four sate-of-the-art continual learning approaches that work with a fixed model with capacity reusing. LwF \cite{LwF} and EWC \cite{EWC} are popular regularization-based approaches for handling catastrophic forgetting, and HAT \cite{HAT} and KAN \cite{KAN} are two masking-based approaches by learning task-specific masks with particular objectives. We also consider the three different reference approaches in continual learning. The results of multi-task  learning where all tasks are jointed learned are also reported as an upper bound for sequential learning.
\begin{itemize}
	\item Individual Networks. It trains an individual network from scratch for each task. This baseline gives the no-reusing solution for minimum forgetting at the cost of large capacity for each task.	
	\item Classify-only. All tasks use the same encoder learned with the data of the first task and only train a task-specific classifier with the data of each task. 
	\item No-masking. It has the same model structure as Classify-only, but all parameters of the encoder are sequentially learned on all tasks. It is a full knowledge sharing approach without any direct control over forgetting. 	
\end{itemize}

\textbf{Network and training.} 
We use the same network architectures and hyperparameter settings as \cite{KAN}, i.e., a single layer GRU for RNN encoder, a fully-connected layer for each classifier and the input embedding with pretrained BERT-base \cite{devlin2018bert}. We set the retention ratio $p$ in TaskDrop to 0.8, 0.5, and 0.6, respectively for high-6, mix-24, and low-6 when comparing with other continual learning approaches. 

Each approach is evaluated in terms of averaged accuracy $A^{\leq t}$ and forgetting ratio $\rho^{\leq t}$ on the testing data of each task after a short-term learning of two tasks and long-term learning of $T$ tasks. $T$ is the total number of tasks of each dataset. Assume $a^{\tau \leq t}$ is the testing accuracy for each task $\tau$ after training $t$ tasks, the averaged accuracy takes average over all the $t$ tasks, i.e., $A^{\leq t}=\frac{1}{t} \sum_{\tau=1}^{t} a^{\tau \leq t}$. Forgetting ratio \cite{HAT} is an adjusted variant of accuracy by comparing to the random and joint learning accuracy.

Since the results are order-dependent, we generate 10 different sequences randomly for each dataset, and report the mean and standard deviations of these 10 sequences for all comparisons. All results are reproducible with settings used here. Codes will be published upon acceptance of the paper.

\begin{figure}
	\centering
	\includegraphics[width=0.45\textwidth]{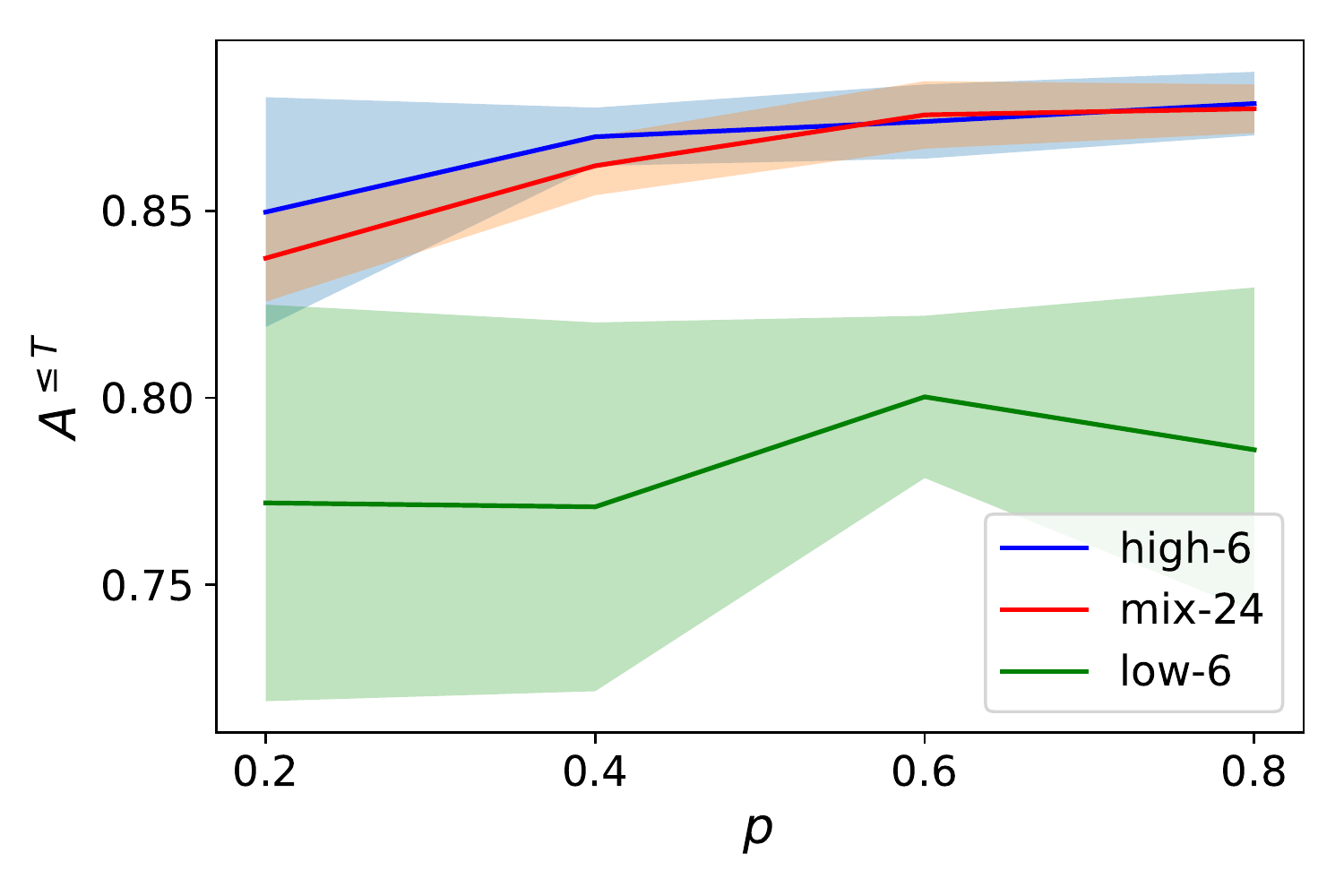}
	\caption{Performance in $A^{\leq T}$ of TaskDrop on three datasets with respect to different retention ratio $p$. }
	\label{fg:mask_rate_6_task}
\end{figure}

\subsection{Comparison with Reference Approaches}
Table \ref{Table:ave_three_ref} compares TaskDrop and four reference approaches on the three datasets. The average of the means over three datasets are also given to show the overall performance over three datasets. The results with respect to two metrics are pretty much consistent, but forgetting ratio is more sensitive than accuracy, i.e., two results with very close accuracy may have a large difference in forgetting ratio. 

It is no surprise that Multi-task gives the best results for all the cases as it jointly learns data of all the tasks. For the rest approaches that have no access to data of previous tasks, TaskDrop and No-masking, the two which encourage more capacity reuse perform better. While the Individual Networks was reported to give the best results on image classification in \cite{PackNet}, it performs much worse than all others for these sentiment datasets. By comparing No-masking's long-term results with its short-term ones, it is seen that results of the three datasets are improved when learning more tasks. 
All the above observations confirm that forgetting is the main issue only for some continual learning problems, but not for sentiment classification.

\begin{figure*}
	\centering
	\includegraphics[width=\textwidth]{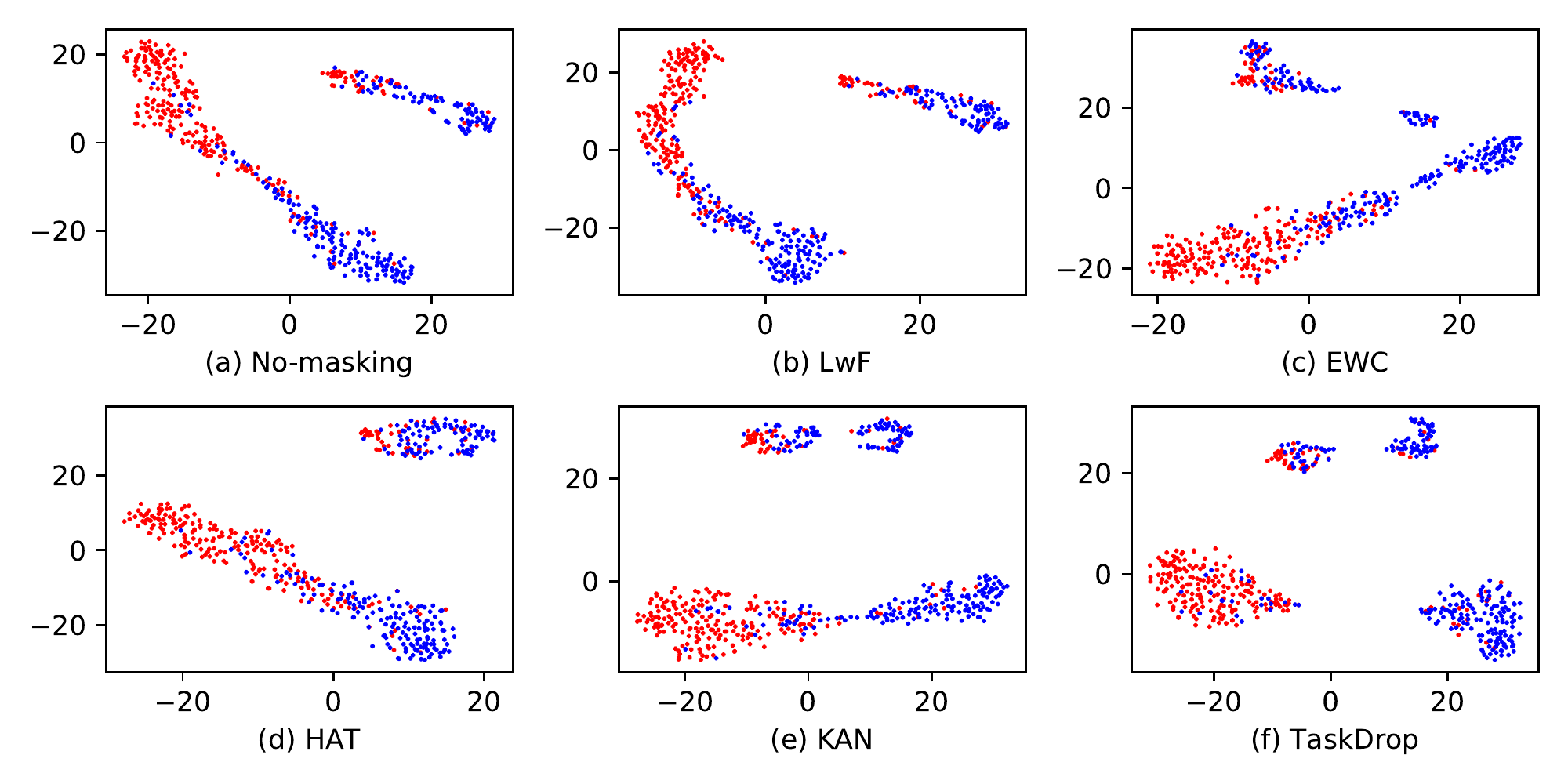}
	\caption{Representations of testing instances in \textit{Instant Video} generated with different approaches, which are trained on other tasks in low-6. Representations are visualized by t-SNE \cite{van2008visualizing}. }
	\label{fg:visulization}
\end{figure*}

Comparing the two top approaches, TaskDrop outperforms No-masking from a long term, and No-masking gives good results on the two datasets with large cross-task relevance. As discussed early, random masking equips TaskDrop with the ability of skip-task transfer, which helps to reduce forgetting. However, the learning period of two tasks is too short to allow any skip-task transfer. Nevertheless, TaskDrop achieves the best results on low-6 with respect to both short and long term learning. Because tasks of this dataset share less knowledge to each other, making full capacity sharing in No-masking less effective. The above observations indicate that even for sentiment classification tasks, randomly masking with a proper ratio still helps to learn a more robust and hence effective model from a relative long-term point of view.

\subsection{Comparison with State-of-the-art Approaches}	
Table \ref{Table:ave_three_SOTA} compares TaskDrop with the four state-of-the-art continual learning approaches in the same way as above. Again, the results of TaskDrop after learning all the $T$ tasks are the best on all three datasets. 

The transfer-focused approach KAN outperforms the other three that dedicated for catastrophic forgetting. 
By comparing to results in Table \ref{Table:ave_three_ref}, we found that  No-Masking beats all of these state-of-the-art ones for most of the cases, except some cases on low-6, showing that the specific ways those approaches formulated for selective knowledge transfer are not suitable for datasets used here.      
We also compare different approaches via visualizing the learned representations in Figure \ref{fg:visulization}, which plots the representations of test instances in the task of \textit{Instant Video} with model trained on other tasks in low-6. It is seen that compact clusters are formed with features learned by TaskDrop. Two pairs of clusters with different class labels which are merged by other approaches are separated with a large margin by TaskDrop.

\begin{figure}
	\centering
	\includegraphics[width=0.6\textwidth]{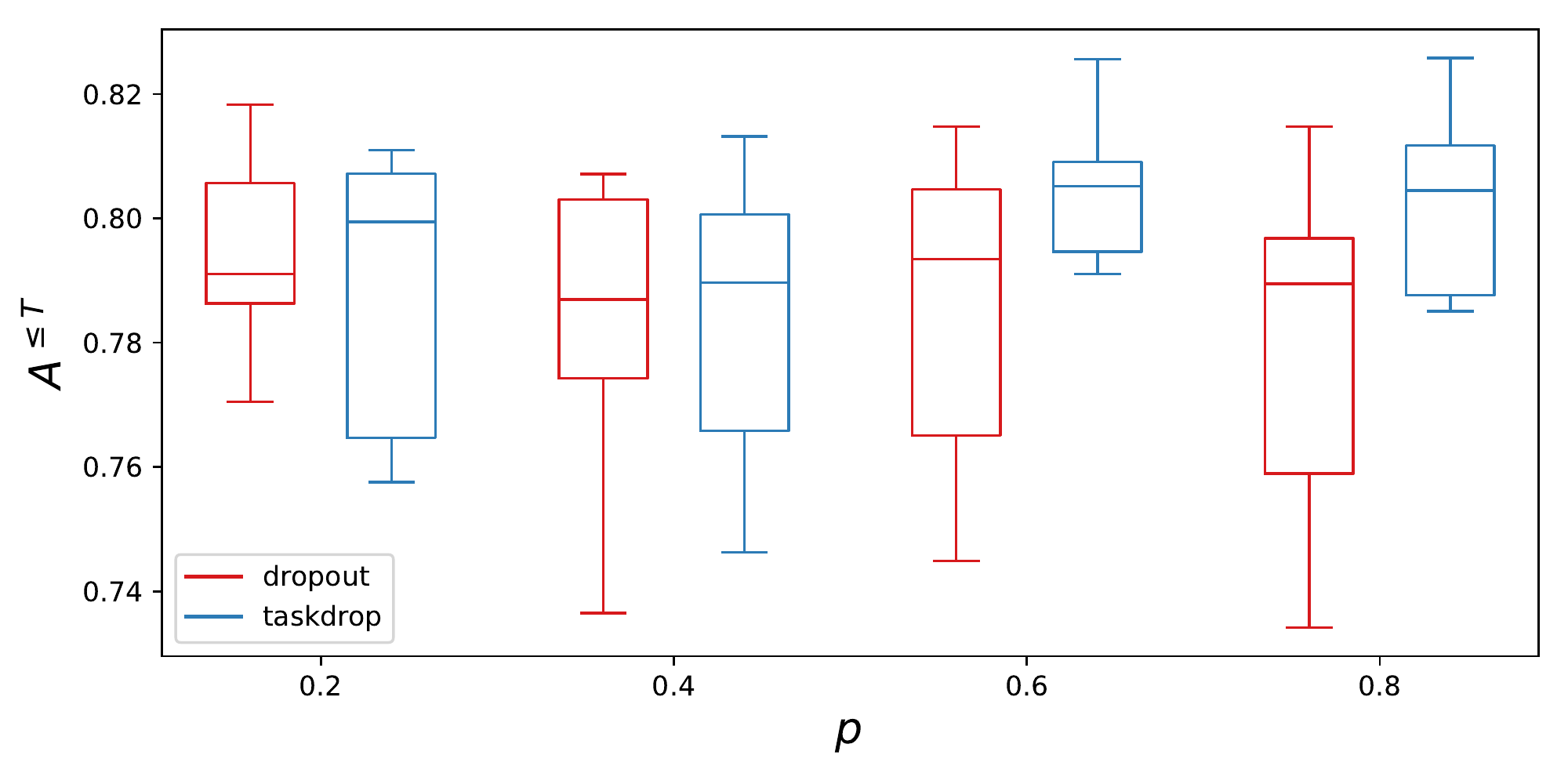}
	\caption{Comparison of TaskDrop and dropout in $A^{\leq T}$ on low-6 with respect to different retention ratio $p$. }
	\label{fg:dropuout}
\end{figure}

\subsection{TaskDrop vs. Dropout}
Now we compare the results of TaskDrop with the standard dropout. As seen from Figure \ref{fg:dropuout} that TaskDrop performs better than dropout when $p$ is not too small, i.e., $p\geq 0.4$. 
This confirms that 
knowing the task boundary during training is important for the continual learning setting, which is not designed in dropout. The results of TaskDrop is slightly worse than dropout for $p=0.2$ as the ensemble of large number of subnetworks makes it less challenging for dropout to learning with a very sparse model. 

\subsection{Retention Ratio}
Finally, we study the impact of the retention ratio $p$. Figure \ref{fg:mask_rate_6_task} plots the results of TaskDrop in $A^{\leq T}$ with respect to different $p$ for the three datasets. 
The three curves have a similar overall tendency, i.e., going up when increasing $p$ from 0.2 and reaching the maximums with $p$ between 0.6 and 0.8, and then slowly going down. Compared to the low-6 curve, the other two are much closer to each other in both shape and vertical position, indicating that low-6 has a more dissimilar nature from the other two datasets. The optimal value of $p$ is smaller for low-6, which is not surprising as tasks of this dataset are less transferable. 
As a too small $p$ causes the sampled encoder too simple to work reasonably, and also results in large standard deviations. Even though, with only 20\% units of each layer, the results of TaskDrop are on par or better than those of Individual Networks. 

\section{Conclusion}
We investigated a random masking-based approach for continual sentiment classification, where tasks are coming sequentially. The proposed approach gives competitive performance in our experiments without introducing extra memory or learning modules, demonstrating the effectiveness of random masking based capacity allocation and reusing for the problem considered in this study.
It will be interesting to further investigate this simple framework on problems from other domains to see whether the proposed random masking mechanism also works well for those problems that have different natures from sentiment classification tasks.

\bibliographystyle{unsrtnat}
\bibliography{references}

\end{document}